\pdfoutput=1

\documentclass[11pt]{article}

\usepackage[final]{acl}

\usepackage{times}
\usepackage{latexsym}
\usepackage{booktabs}
\usepackage{multirow, makecell}
\usepackage{tikz}
\usetikzlibrary{shapes,backgrounds}

\usepackage{centernot}
\usepackage{marvosym}
\usepackage{etoolbox}
\usepackage{cuted}

\AfterEndEnvironment{strip}{\leavevmode}

\usepackage{fdsymbol}
\usepackage[T1]{fontenc}
\usepackage[utf8]{inputenc}
\usepackage{microtype}
\usepackage{inconsolata}

\usepackage{graphicx}
\usepackage{tabularx}
\usepackage{algpseudocode}
\usepackage{microtype}
\usepackage{subcaption}
\usepackage{algorithm}
\usepackage{algpseudocode}
\usepackage{multirow}
\usepackage{multicol}
\usepackage{xcolor,colortbl}
\usepackage{amsmath}
\DeclareMathOperator*{\argmax}{\textit{\textbf{argmax}}}
\usepackage{xspace}

\definecolor{dkblue}{rgb}{0,0,0.5}

\newcommand{\bfund}[1]{\textbf{#1}}

\newcommand{\doublespace}{\space\space}

\title{Emergent Abilities of Large Language Models\\under Continued Pretraining for Language Adaptation}

\author{Ahmed Elhady$^{1}$ \quad Eneko Agirre$^{1}$ \quad Mikel Artetxe$^{1,2}$ \\
$^{1}$HiTZ Center, University of the Basque Country (UPV/EHU) \qquad $^{2}$Reka AI \\
\texttt{\{ahmed.salemmohamed,e.agirre,mikel.artetxe\}@ehu.eus} }

\begin{document}
\maketitle

\begin{abstract}
  Continued pretraining (CPT) is a popular approach to adapt existing large language models (LLMs) to new languages. When doing so, it is common practice to include a portion of English data in the mixture, but its role has not been carefully studied to date. In this work, we show that including English does not impact validation perplexity, yet it is critical for the emergence of downstream capabilities in the target language. We introduce a language-agnostic benchmark for in-context learning (ICL), which reveals catastrophic forgetting early on CPT when English is not included. This in turn damages the ability of the model to generalize to downstream prompts in the target language as measured by perplexity, even if it does not manifest in terms of accuracy until later in training, and can be tied to a big shift in the model parameters. Based on these insights, we introduce curriculum learning and exponential moving average (EMA) of weights as effective alternatives to mitigate the need for English. All in all, our work sheds light into the dynamics by which emergent abilities arise when doing CPT for language adaptation, and can serve as a foundation to design more effective methods in the future. 

\end{abstract}


\begin{figure}[t!]
\begin{subfigure}[b]{1\columnwidth} 
    \includegraphics[width=1\columnwidth]{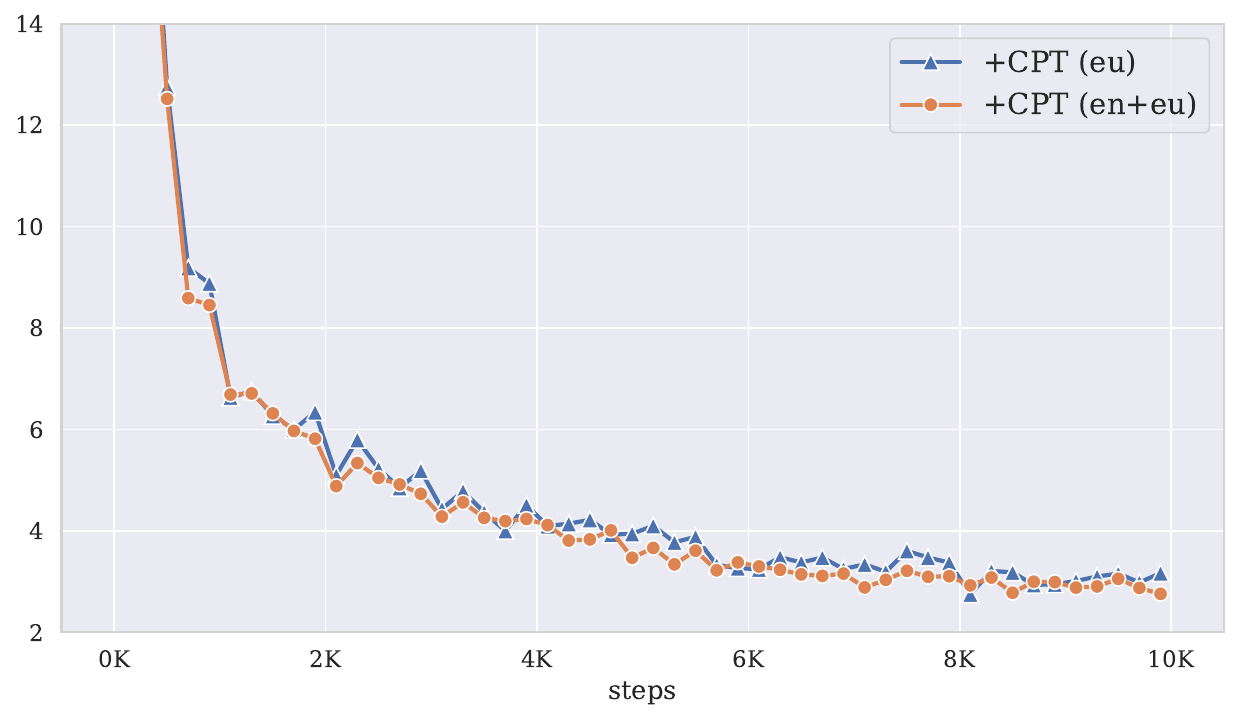}
    \caption{Validation perplexity (eu)} \label{subfig:ppl}
\end{subfigure}
\begin{subfigure}[b]{1\columnwidth}
   \includegraphics[width=1\columnwidth]{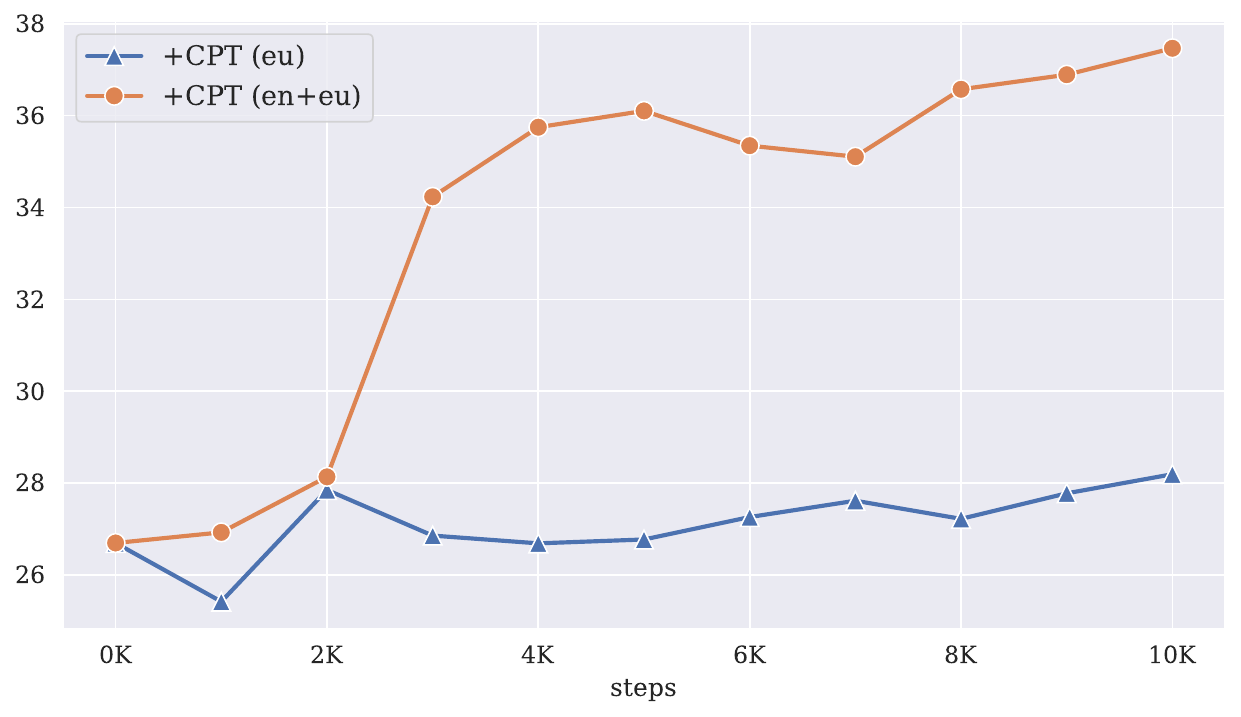}
   \caption{Downstream accuracy (eu)} \label{subfig:downstream}
\end{subfigure}
\caption{\textbf{Continued pretraining of Llama 2 7B on Basque data with and without including English data.} Both models exhibit similar validation perplexity on Basque (top), yet the variant including English significantly outperforms on downstream tasks (bottom).}
    \label{fig:prelm-exp}
\end{figure}

\section{Introduction}

Despite achieving remarkable results in multilingual tasks like machine
translation \cite{zhu2024multilingualmachinetranslationlarge}, existing large
language models (LLMs) are notoriously English-centric, and their performance
has been reported to drop significantly in less-resourced languages \cite{mgpt,
  yong2023bloom1addinglanguagesupport,
  workshop2023bloom176bparameteropenaccessmultilingual, talat-etal-2022-reap}.
This has motivated a large body of work to extend existing LLMs to new
languages through continued pretraining (CPT)
\cite{gogoulou2024continuallearninglanguageshift,
  etxaniz2024latxaopenlanguagemodel,
  luukkonen-etal-2023-fingpt, yong2023bloom1addinglanguagesupport}. In its most
basic form, CPT uses an existing LLM as initialization and fine-tunes all
parameters on next-token prediction over a monolingual corpus in the target
language.

Nevertheless, vanilla CPT is rarely used in practice. Instead, there are two
techniques that are broadly used in the literature: (i) mixing target language
data with English or other languages in the original mixture
\cite{etxaniz2024latxaopenlanguagemodel,
  gogoulou2024continuallearninglanguageshift}, and (ii) using LORA
\cite{hu2021loralowrankadaptationlarge} or other parameter-efficient
fine-tuning methods \cite{cui2024efficienteffectivetextencoding,
  yong2023bloom1addinglanguagesupport}. There are inherent advantages to these
techniques that have typically been used to justify their adoption, such as
preserving English performance when including data in this language
\cite{fujii2024continualpretrainingcrosslingualllm,
  cui2024efficienteffectivetextencoding}, or reducing memory requirements when
performing parameter-efficient fine-tuning
\cite{hu2021loralowrankadaptationlarge}. Perhaps more intriguingly, there have
also been isolated reports of these two techniques improving performance in the
target language \cite{ji2024emma500enhancingmassivelymultilingual,
  etxaniz2024latxaopenlanguagemodel}.

Through systematic experiments, we corroborate that including English data when
doing CPT is critical to obtain strong in-context learning (ICL) performance in
the target language (\S\ref{subsec:final-performance}). For instance, we obtain
considerably better results on Basque downstream tasks when performing CPT of
Llama 2 on a mixture of Basque and English data, as opposed to Basque alone
(Figure \ref{subfig:downstream}). But, to our surprise, we find that both
mixtures perform at par in terms of Basque perplexity (Figure
\ref{subfig:ppl}).
We find this to be counterintuitive: both models do equally well in terms of
the pretraining objective in the target language,\footnote{Note that perplexity
  is the exponential of cross-entropy, which is used as the loss.} yet downstream
capabilities only emerge in one of the variants, challenging prior observations
in monolingual settings that models with a similar perplexity tend to obtain
similar performance in downstream tasks
\cite{du2024understandingemergentabilitieslanguage,
  xia2023trainingtrajectorieslanguagemodels}.

We present an empirical study of the training dynamics that lead to this
behavior. We introduce Copain, a new benchmark to evaluate ICL in a
language-agnostic manner (\S\ref{sec:copain}), which reveals that CPT without
English suffers from a catastrophic forgetting of its ICL capabilities in the
first few steps of training (\S\ref{subsec:learning-curves}). We further show
that the ability of this variant to generalize to downstream prompts gets
severely damaged at this exact same period as measured by perplexity, even if
it does not manifest in terms of accuracy until much later in training
(\S\ref{subsec:generalization-behavior}). Finally, we show that this behavior
can be tied to a strong shift in the model parameters when English is not
included in the CPT mixture (\S\ref{subsec:param-shift}).

Based on these insights, we explore two alternative approaches that mitigate
the need for English. First, we show that including English in the first 10\%
training steps in a curriculum learning fashion is sufficient, confirming that
the critical period is concentrated in the first stage of CPT
(\S\ref{subsec:curriculum}). Second, we eliminate the need for English by
applying the exponential moving average (EMA) of weights, which acts as a
regularizer to limit the parameter shift (\S\ref{subsec:ema}).

All in all, our work sheds light into the dynamics that condition the emergence
of downstream abilities when doing CPT for language adaptation, which transcend
what is directly observable by inspecting the training loss. We validate our
findings by designing two CPT variants that mitigate the need for English, and
we hope that our analysis can serve as a foundation to design more effective
language adaptation methods in the future.

\section{Experimental Setup}
\label{sec:experimental-setup}

Our focus is to analyze the CPT of English-centric LLMs for language
adaptation. To that end, we conduct separate experiments on 3 diverse target
languages: Basque, Arabic and Indonesian. We next detail our experimental
settings.

\paragraph{Base models.} We use Llama 2 7B \cite{touvron2023llama2openfoundation} as the base model for
most of our experiments. We base this choice on the original pre-training of
Llama 2 being notoriously English-centric,\footnote{According to the authors,
  their pretraining dataset included 0.03\% of Indonesian and less than 0.005\%
  of Basque and Arabic.} making it a good fit to study CPT under language shift.
To understand the impact of scale and different base models, we run additional
experiments on Basque using Llama 2 13B, Llama 3.1 8B
\cite{dubey2024llama3herdmodels} and Gemma 2 9B
\cite{gemmateam2024gemma2improvingopen}. Since the scripts of the languages we
use are already supported in llama 2, we do not change the vocabulary. This is
also to keep the experiments controlled and not add another dimension for the
vocabulary.


\paragraph{Training data.} For Basque, we use the Latxa corpus \citep{etxaniz2024latxaopenlanguagemodel},
which consists of 4.7B tokens of high-quality Basque text. For Arabic and
Indonesian, we randomly sample documents from their respective CulturaX corpus
\citep{nguyen2023culturaxcleanedenormousmultilingual}. We ensure all languages
have token count parity (4.5$\sim$4.7B tokens per language). When including
English in the CPT mixture, we use a random sample of 500k English documents
from the Pile \citep{gao2020pile800gbdatasetdiverse}. For all languages, the
English data accounts for 20\% of the total CPT tokens.

\paragraph{Hyperparameters.} All models are continued pretrained for 10k steps on $4 \times 8$ A100 GPUs.
The learning rate is set to 1E-04 with cosine scheduling and a 10\% warm-up
ratio. The maximum sequence length is set to 4096 and the effective batch size
to 256. These hyperparameters were chosen in accordance with
\citet{etxaniz2024latxaopenlanguagemodel}; we did not observe any significant
impact when varying them in our early experiments (see Appendix
~\ref{app:prel-exp}).

\paragraph{Evaluation.} For all models, we report the perplexity on the validation split of their
respective data. We assess the performance on downstream tasks using
multiple-choice benchmarks. We choose natively curated benchmarks in their
respective languages to avoid translation artifacts
\cite{artetxe-etal-2020-translation}. In addition, we report results on MGSM-eu
\cite{baucells-etal-2025-iberobench}, a generative maths dataset that was human
translated from English, in Appendix~\ref{res:mgsm-eu}, where we observe
similar trends.

For Arabic and Indonesian, we report accuracy on ArabicMMLU
\citep{koto2024arabicmmlu} and IndoMMLU
\citep{koto2023largelanguagemodelspass}, respectively. Both benchmarks consist
of 5 sub-tasks measuring language proficiency, reasoning ability, and cultural
knowledge of their respective language. For Basque, we report average accuracy
across EusTrivia, EusProficiency, EusExams and EusReading
\citep{etxaniz2024latxaopenlanguagemodel}. All benchmarks use multiple choice
prompting \cite{robinson2023leveraginglargelanguagemodels} with 5-shot
examples, except for EusReading which uses 1-shot (see Appendix~\ref{app:mcp}
for details). In addition, we report accuracy in Copain---a new
language-agnostic ICL benchmark we introduce in \S\ref{sec:copain}---for all
languages.

\section{Copain: A Language-Agnostic ICL Benchmark}
\label{sec:copain}

Multiple-choice benchmarks require both (i) a good level of ICL (so the LLM
generates an answer in the expected format based on the few-shot
demonstrations), and (ii) knowledge of the relevant task in the target
language. Intuitively, (i) is not tied to any specific language, so the initial
models should already be capable on it, while (ii) should improve as we perform
CPT in the target language. However, the fact that these two aspects are
conflated in downstream metrics makes it hard to understand why certain
variants underperform others. Are they less effective at learning the target
language? Or do they become weaker at ICL?

\begin{table}[t!]
\renewcommand{\arraystretch}{0.75} 
\small
\resizebox{1\columnwidth}{!}{

\begin{tabular}{ll}
\toprule
\textbf{Task}&\textbf{Example Prompt}\\
\midrule
\multirow{4}{*}{\textbf{Max. integer in list}}&85,\doublespace24,\doublespace63$\colon$\doublespace85\\
&29,\doublespace47,\doublespace79$\colon$\doublespace79\\
&59,\doublespace77,\doublespace41$\colon$\doublespace77\\
&19,\doublespace81,\doublespace88$\colon$\\
\midrule
\multirow{4}{*}{\textbf{Min. integer in list}}&85,\doublespace24,\doublespace63$\colon$\doublespace24\\
&29,\doublespace47,\doublespace79$\colon$\doublespace29\\
&59,\doublespace77,\doublespace41$\colon$\doublespace41\\
&19,\doublespace81,\doublespace88$\colon$\\
\midrule
\multirow{4}{*}{\textbf{Median integer in list}}&85,\doublespace24,\doublespace63$\colon$\doublespace63\\
&29,\doublespace47,\doublespace79$\colon$\doublespace47\\
&59,\doublespace77,\doublespace41$\colon$\doublespace59\\
&19,\doublespace81,\doublespace88$\colon$\\
\midrule
\multirow{4}{*}{\textbf{Even integer among odd list}}&21,\doublespace71,\doublespace68,\doublespace95$\colon$\doublespace68\\
&25,\doublespace35,\doublespace58,\doublespace83$\colon$\doublespace58\\
&92,\doublespace71,\doublespace61,\doublespace29$\colon$\doublespace92\\
&97,\doublespace66,\doublespace\doublespace1,\doublespace\doublespace3$\colon$ \\
\midrule
\multirow{4}{*}{\textbf{Odd integer among even list}}&24,\doublespace76,\doublespace60,\doublespace51$\colon$\doublespace51\\
&83,\doublespace52,\doublespace22,\doublespace52$\colon$\doublespace83\\
&32,\doublespace68,\doublespace10,\doublespace79$\colon$\doublespace79\\
&64,\doublespace87,\doublespace\doublespace0,\doublespace28$\colon$ \\
\midrule
\multirow{4}{*}{\textbf{First character alphabetically}}&w,\doublespace y,\doublespace\space a$\colon$\doublespace a\\
&b,\doublespace m,\doublespace k$\colon$\space b\\
&v,\doublespace\space h,\doublespace p$\colon$\space h\\
&y,\doublespace e,\doublespace\space p$\colon$ \\
\midrule
\multirow{4}{*}{\textbf{Last character alphabetically}}&w,\doublespace y,\doublespace\space a$\colon$\doublespace y\\
&b,\doublespace m,\doublespace k$\colon$\space m\\
&v,\doublespace\space h,\doublespace p$\colon$\doublespace v\\
&y,\doublespace\space e,\doublespace p$\colon$ \\
\bottomrule
\end{tabular}
}
\caption{\textbf{Example Copain tasks using 3-shot demonstrations.} The model's predictions are evaluated using exact match accuracy.}
\label{tab:copain}
\end{table}

So as to evaluate ICL in a language-agnostic manner, we introduce the
\bfund{Co}ntextual \bfund{pa}ttern \bfund{in}ference (Copain) benchmark. As
shown in Table~\ref{tab:copain}, the input of the task is a list of either
numbers or characters, and the model needs to output the element in the list
that meets certain criterion. However, there is no natural language instruction
in the prompt, so the model needs to infer the task from the few-shot
demonstrations.


The benchmark comprises 7 tasks. Each task consists of 150 examples, totaling
1050. The tasks are to identify: (i) the minimum/maximum/median number in a
list of 3 integers, (ii) the even/odd number in a list of 4 integers, and (iii)
the alphabetically first/last character in a list of 3 Latin characters. We use
exact-match accuracy as the evaluation metric. 

\begin{table}[t!]
\centering
\small
\begin{tabular}{lrrr}
\toprule
&\textbf{PPL}&\textbf{Dwn.} &\textbf{Cop.}\\
\rowcolor{gray!25} 
\bottomrule
\multicolumn{4}{l}{\textbf{Basque (eu)}}\\
\toprule
Llama 2 (7B)&23.64&27.43&\bfund{44.67}\\
\quad + CPT (eu+en)&\bfund{3.35}&\bfund{34.14}&43.43\\
\quad + CPT (eu) &3.58&28.89&20.12\\
\midrule
Llama 2 (13B)&13.66&29.52&\bfund{49.23}\\
\quad + CPT (eu+en)&2.82&\bfund{42.52}&47.80\\
\quad + CPT (eu)&\bfund{2.79}&35.20&29.43\\
\midrule
Llama 3.1 (8B)&2.18&42.31&41.32\\
\quad + CPT (eu+en)&\bfund{1.73}&\bfund{55.75}&\bfund{42.04}\\
\quad + CPT (eu)&1.82&54.84&41.19\\
\midrule
Gemma 2 (9B)&2.28&42.22&\bfund{51.90}\\
\quad + CPT (eu+en)&1.52&\bfund{49.39}&50.23\\
\quad + CPT (eu)&\bfund{1.48}&45.95&43.59\\
\bottomrule
\rowcolor{gray!25} 
\multicolumn{4}{l}{\textbf{Arabic (ar)}}\\
\toprule
Llama 2 (7B)&4.36&32.45&\bfund{44.67}\\
\quad + CPT (ar+en)&\bfund{2.09}&\bfund{34.34}&32.60\\
\quad + CPT (ar)&2.12&32.67&23.80\\
\bottomrule
\rowcolor{gray!25} 
\multicolumn{4}{l}{\textbf{Indonesian (id)}}\\
\toprule
Llama 2 (7B)&6.27&26.65&\bfund{44.67}\\
\quad + CPT (id+en)&3.25&\bfund{30.79}&30.79\\
\quad + CPT (id)&\bfund{3.05}&26.92&27.34\\
\bottomrule
\end{tabular}
\caption{\textbf{Main results for each base model and its continued pretraining with and without English.} We report validation perplexity in the target language (\textit{PPL}), average downstream accuracy in the target language (\textit{Dwn}; see \S\ref{sec:experimental-setup} for details), and Copain accuracy (\textit{Cop}).}
\label{tab:final-performance}
\end{table}

\section{The Impact of English in CPT}
\label{sec:imapct-eng}

In this section, we study the impact of English when doing CPT for language
adaptation, covering the final performance at the end of CPT
(\S\ref{subsec:final-performance}), the learning trajectory
(\S\ref{subsec:learning-curves}), the generalization behavior as measured by
perplexity (\S\ref{subsec:generalization-behavior}), and the parameter shift
(\S\ref{subsec:param-shift}).

\subsection{Final performance}
\label{subsec:final-performance}

We start by analyzing the final performance of the models at the end of CPT. As
shown in Table \ref{tab:final-performance}, CPT brings big gains over all base
models in terms of target language perplexity. Both CPT variants perform at
par: the one with English wins in 3 instances and the one without wins in the
remaining 3, but the differences are small in all cases. This suggests that
including English does not directly help language modeling in the target
language, at least in terms of the pretraining objective itself.

However, we do observe notable differences in downstream performance. Just as
with perplexity, all CPT models outperform their corresponding base model. But,
in this case, the variant including English obtains considerably better
results, beating the variant not including English in all cases. The weaker the
base model is in a given language, the more helpful including English tends to
be, with a difference over 7 points for Llama 2 13B in Basque.

Results in Copain show a similar trend: the CPT variant with English
outperforms the one without in all cases. The differences tend to be large
(e.g., around 20 points for Llama 2 in Basque), although they greatly vary
across languages and base models. However, different from perplexity and
downstream tasks, it is the initial model that obtains the best results in most
cases (5 out of 6). This suggests that doing CPT for language adaptation tends
to harm the ICL capabilities of LLMs, and including English helps mitigate
this.

All in all, our results show that including English during CPT leads to
considerably better downstream performance. However, this difference does not
manifest in perplexity, and can instead be attributed to a better preservation
of the ICL capabilities of the original model.



\begin{figure}[t!]
\includegraphics[width=1.\columnwidth]{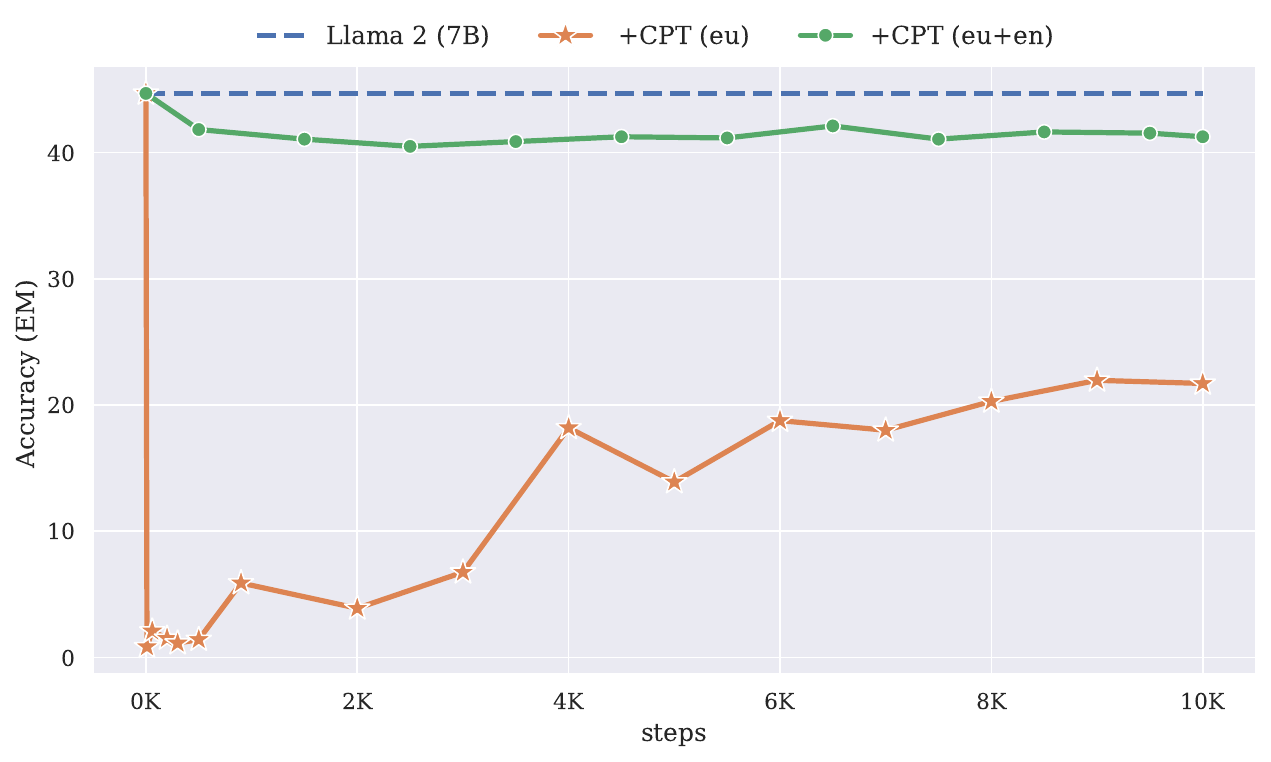}
\caption{\textbf{Copain results for Llama 2 7B.} Including English during CPT retains over 94\% of the original performance, while not including it results in catastrophic forgetting followed by a slow partial recovery. }
\label{fig:copain}
\end{figure}

\subsection{Learning Trajectory}
\label{subsec:learning-curves}

Our results so far were limited to the final performance of the models. In this
section, we will look at how their behavior evolves throughout CPT. To that
end, we will focus on Llama 2 7B in Basque, for which we have previously
observed one of the biggest differences in final performance.

As shown in Figure \ref{subfig:ppl}, the learning curve for perplexity looks
very similar regardless of whether English is included or not. In contrast,
downstream performance shows an emergent behavior when including English, with
a sudden improvement of 8 points between steps 2k and 4k, while it never takes
off when English is not included (Figure \ref{subfig:downstream}). This
challenges prior findings that models with a similar perplexity obtain a
similar downstream performance, with certain abilities emerging when perplexity
falls below a certain threshold
\citep{xia2023trainingtrajectorieslanguagemodels,du2024understandingemergentabilitieslanguage}.
While those studies focused on monolingual pretraining, we show that this
behavior does not hold more broadly when doing CPT for language adaptation.

Figure \ref{fig:copain} shows that both CPT variants behave differently in
Copain too. When English is not included, we observe catastrophic forgetting
early on training, with performance plummeting to nearly zero in the first few
steps. This is followed by a slow improvement throughout the rest of training,
which is far from recovering the full performance of the original model. The
CPT variant with English suffers a more progressive degradation in the first 2k
steps, but it is very mild in comparison, and performance remains nearly
constant after that.

Based on these results, we hypothesize that there is a critical period at the
beginning of CPT, where the strong distribution shift from switching to a new
language can result in a catastrophic forgetting of the ICL capabilities of the
model. This would in turn prevent the emergence of downstream capabilities
later on, even if not directly impacting the training objective as reflected by
the validation perplexity. Including English data in the mixture would
alleviate this distribution shift, mitigating the catastrophic forgetting.





\subsection{Generalization Behavior}
\label{subsec:generalization-behavior}

We have so far established that including English data in CPT is critical for
good downstream and ICL performance, despite not having an impact on validation
perplexity. In other words, the two CPT variants perform similarly when
evaluated in the training distribution,\footnote{Our training and validation
  datasets are obtained by taking two random splits of the original corpora, and
  they thus come from the same distribution.} but generalize differently to
few-shot tasks that are out of this distribution. However, the two aspects are
evaluated using different metrics (perplexity vs. accuracy). The former is a
function of the training loss, but the latter is not directly tied to it,
making it difficult to understand the nature of this different generalization
behavior. To overcome this, we next formulate downstream tasks as conditional
text generation, and use perplexity to evaluate different models on it.




Given a set of few-shot demonstrations \textit{C} and a question \textit{Q},
the model predicts the probability of each answer label $A$ conditioned on the
prefix prompt consisting of $T$ tokens:
\begin{equation}
  \renewcommand{\arraystretch}{0.4} %
  \label{eqn:icl-prob}
  \textsc{P}(A|C,Q) = \prod_{t=1}^{T} p(A_t|C,Q<T)
\end{equation}
The perplexity of the answers aggregated over the entire test set of $N$ examples can be computed as follows:
\begin{equation}
  \renewcommand{\arraystretch}{0.4} %
  \label{eqn:ppl}
  \textsc{PPL}(A| Q,C) = 2^{-\frac{1}{N} \sum_{i=1}^{N} \log P(a_i |c_i,q_i)}
\end{equation}

\begin{figure}[t!]
        \includegraphics[width=1\columnwidth]{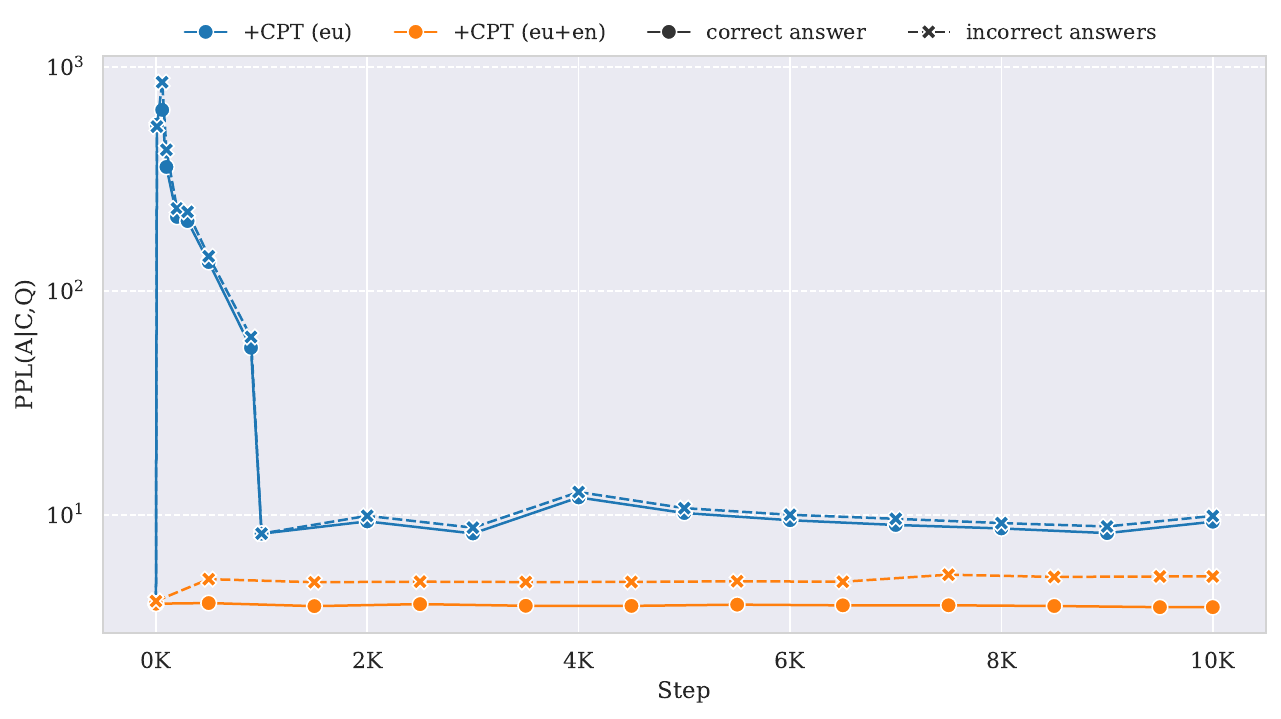}
    \caption{\textbf{Perplexity of choice labels on Basque downstream tasks for Llama 2 7B.} The variant without English experiences a spike in perplexity simultaneous with the drop in ICL (Figure~\ref{fig:copain}). \textit{PPL} of incorrect labels are averaged.}
    \label{fig:ppl-a-qc}
\end{figure}

We separately compute the perplexity of the correct and incorrect answers.
Intuitively, we want the gap between the two to be as high as possible, as
strong models should assign a higher probability to correct answers than to
incorrect answers. As shown in Figure \ref{fig:ppl-a-qc}, CPT with English is
effective at achieving this: the perplexity of incorrect answers increases
early on training, while the perplexity of correct answers remains constant or
even slightly goes down. When English data is not included, the perplexity of
both answers spikes in the first few steps of CPT. Even if it tends to
stabilize later on, it stays high compared to the variant with English.
Moreover, the gap between the correct and the incorrect answers is much
smaller.

In conclusion, when it comes to their pure language modeling performance, both
CPT variants behave similarly in the training distribution. But, when not
including English, the ability to generalize to multiple choice prompts that
are out of this distribution gets severely damaged after the first few steps.
Even if the difference in downstream accuracy becomes prominent later on
training (around step 3k in Figure \ref{subfig:downstream}), this shows that
the real damage happens much earlier, in line with the drop observed for Copain
in \S\ref{subsec:learning-curves}.



\subsection{Parameters Shift}
\label{subsec:param-shift}

Our experiments so far have shown that excluding English from the mixture
causes catastrophic forgetting of some emergent abilities in the first few
steps of CPT. In this section, we analyze the underlying training dynamics that
cause this behavior. To that end, we measure how much the model parameters
change with respect to their initial value as training progresses.

\begin{figure}[t!]
\includegraphics[width=1\columnwidth]{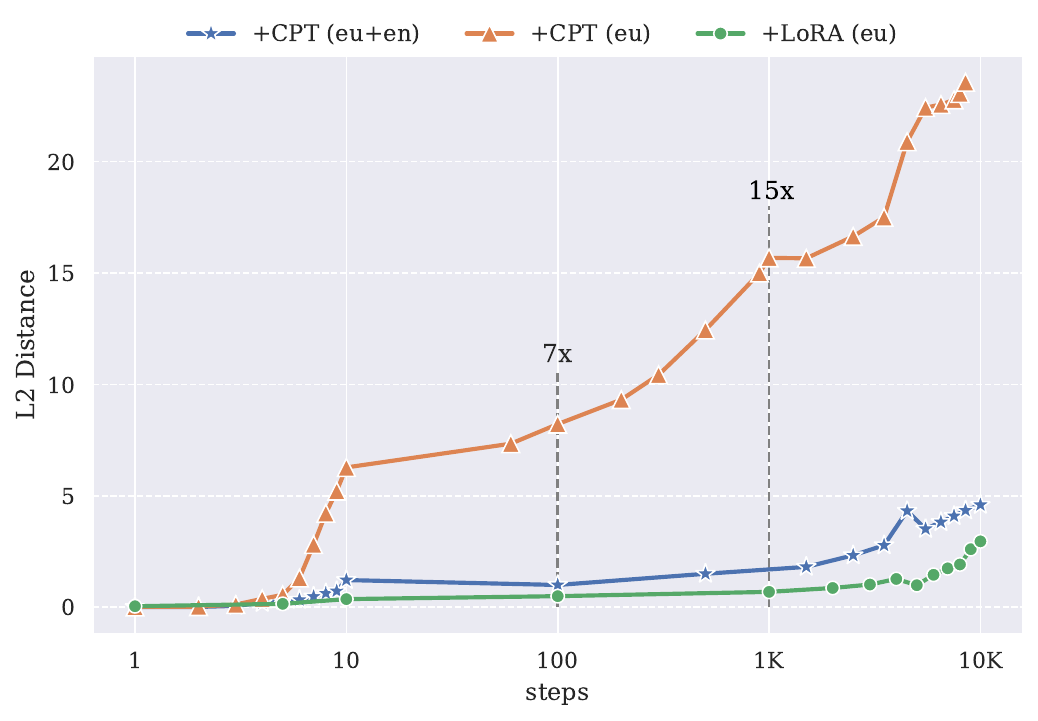}
\caption{\textbf{Average layer-wise L2 distance of model parameters} from the initial Llama 2 7B model throughout full-parameter CPT and using LoRA.  The CPT variant without English data rapidly diverges from the initial weights during the first 1k steps. The divergence is slowed for the rest of the training steps. }
\label{fig:l2dist}
\end{figure}

As shown in Figure~\ref{fig:l2dist}, the variant without English experiences a
stronger parameter shift. The shift rapidly builds up during the first few
steps: at the 100th step, the cumulative L2 distance is 7x higher for the
variant without English, reaching 15x by the 1000th step. In contrast, CPT with
English shows a more steady and regularized parameter change. In fact, the
variant without English undergoes a bigger change in the first 10 steps than
the variant with English during the entire course of training.

We further run an experiment using LoRA with Basque data only.\footnote{We set
  the rank to 512 following \citet{talla2024neutralresiduesrevisitingadapters},
  which corresponds to 20\% additional trainable parameters.} We find that LoRA
has a similar effect to including English, with an even smaller shift in model
parameters.\footnote{LoRA reparameterizes the model weights as $W=W_0 + AB$,
  where $W_0$ are the initial model weights (which are kept frozen), and $A$ and
  $B$ are learnable, low-rank matrices. This way, even if the number of learnable
  parameters is reduced, one can still measure the parameter shift from $W_0$ to
  $W$ in a way that is comparable to full fine-tuning.} As reported in
Appendix~\ref{app:lora}, this approach is quite effective at preserving ICL
performance, but barely improves over the initial model on downstream tasks.
This suggests that overly constraining the parameter shift can hinder learning
the target language, while giving too much flexibility can cause catastrophic
forgetting of ICL capabilities. Given its effectiveness at reducing the
parameter shift, this can explain why, with optimized hyperparameters, prior
work has found LoRA to outperform full-parameter CPT in low-resource languages
when English is not included
\citep{ji2024emma500enhancingmassivelymultilingual,
  yong2023bloom1addinglanguagesupport}.

\section{Alternatives to Including English Data}

Our analysis in \S\ref{sec:imapct-eng} shows that (i) there is a critical
period early on CPT where catastrophic forgetting occurs if not including
English, and (ii) this phenomenon can be tied to a strong shift in the model's
parameters. Based on these insights, we next explore two alternative CPT
techniques that achieve the same effect. 


\begin{table}[t!]
\small
\centering
\begin{tabular}{lrrr}
\toprule
&\textbf{PPL}&\textbf{Dwn.} &\textbf{Cop.}\\
\rowcolor{gray!25} 
\bottomrule
\multicolumn{4}{l}{\textbf{Basque (eu)}}\\
\toprule
Llama 2 (7B)&23.64&27.43&\bfund{44.67}\\
{ \quad + CPT (full)}&3.35&34.14&43.43\\
{ \quad + CPT (curr)}&\bfund{3.08}&\bfund{35.12}&42.94\\
\midrule
Llama 2 (13B)&13.66&29.52&\bfund{49.23}\\
{ \quad + CPT (full)}&2.82&\bfund{42.52}&47.80\\
{ \quad + CPT (curr)}&\bfund{2.65}&42.42&46.33\\
\bottomrule
\rowcolor{gray!25} 
\multicolumn{4}{l}{\textbf{Arabic (ar)}}\\
\toprule
Llama 2 (7B)&4.36&32.45&\bfund{44.67}\\
{ \quad + CPT (full)}&2.09&34.34&32.60\\
{ \quad + CPT (curr)}&\bfund{2.00}&\bfund{34.53}&39.66\\
\bottomrule
\rowcolor{gray!25} 
\multicolumn{4}{l}{\textbf{Indonesian (id)}}\\
\toprule
Llama 2 (7B)&6.27&26.65&\bfund{44.67}\\
{ \quad + CPT (full)}&3.25&\bfund{30.79}&30.79\\
{ \quad + CPT (curr)}&\bfund{3.14}&29.09&31.03\\
\bottomrule
\end{tabular}
\caption{\textbf{Results with English data added for all training steps (\textit{full}) and for the first 10\% steps (\textit{curr}).} We report validation perplexity (\textit{PPL}), average downstream accuracy (\textit{Dwn}) and Copain accuracy (\textit{Cop}).}
\label{tab:curr}
\end{table}

\subsection{Curriculum Learning}
\label{subsec:curriculum}

Even if the difference in downstream accuracy arises around step 3k (Figure
\ref{subfig:downstream}), our results on ICL capabilities (Figure
\ref{fig:copain}), perplexity of downstream choice labels (Figure
\ref{fig:ppl-a-qc}) and parameter shift (Figure \ref{fig:l2dist}) show that the
divergence originates much earlier. This can presumably be attributed to the
strong distribution shift when changing the training language in CPT, and
including English would serve to mitigate this. But what if we make the shift
more gradual in a curriculum learning fashion? Is English really needed after
the initial critical period?

To answer this, we experiment with including English during the first 1k steps,
and omitting it thereafter. As shown in Table \ref{tab:curr}, models trained
with this approach (\textit{curr}) exhibit similar performance compared to
those where English is included throughout the entirety of CPT (\textit{full}).
More concretely, each variant wins in half of the cases for both downstream and
Copain accuracy. Interestingly, the curriculum approach obtains the best
perplexity results in all cases, which we speculate could be attributed to the
additional training budget in the target language from omitting English. In any
case, the differences are small in all cases.

All in all, these results corroborate that the role of English is to provide a
smoother transition to the target language distribution. In line with our
results in Table \ref{tab:final-performance}, this also explains why certain
models like Llama 3.1 and Gemma 2 benefit less from including English: those
initial models are already decent at modeling the target language distribution
(as reflected by their lower validation perplexity), alleviating the
distribution shift in CPT and making the smoother transition from including
English less necessary.



\begin{table}[t!]
\centering
\small
\begin{tabular}{lrcr}
\toprule
&\textbf{PPL}&\textbf{Dwn.} &\textbf{Cop}\\
\rowcolor{gray!25} 
\bottomrule
\multicolumn{4}{l}{\textbf{Basque (eu)}}\\
\toprule
Llama 2 (7B)&23.64&27.43&\bfund{44.67}\\
{ \quad + CPT (eu+en)}&3.35 &34.14&43.43\\
{ \quad + CPT w/ EMA (eu)}&\bfund{2.98}&\bfund{34.89}&42.66\\
\midrule
Llama 2 (13B)&13.66&29.52&\bfund{49.23}\\
{ \quad + CPT (eu+en)}&2.82&\bfund{42.52}&47.80\\
{ \quad + CPT w/ EMA (eu)}&\bfund{2.71}&41.39&42.99\\
\bottomrule
\rowcolor{gray!25} 
\multicolumn{4}{l}{\textbf{Arabic (ar)}}\\
\toprule
Llama 2 (7B)&4.36&32.45&\bfund{44.67}\\
{ \quad + CPT (ar+en)}&2.09&\bfund{34.34}&32.60\\
{ \quad + CPT w/ EMA (ar)}&\bfund{2.03}&33.36&42.76\\
\bottomrule
\rowcolor{gray!25} 
\multicolumn{4}{l}{\textbf{Indonesian (id)}}\\
\toprule
Llama 2 (7B)&6.27&26.65&\bfund{44.67}\\
{ \quad + CPT (id+en)}&3.25&\bfund{30.79}&30.79\\
{ \quad + CPT w/ EMA (id)}&\bfund{2.97}&29.11&33.34\\
\bottomrule
\end{tabular}
\caption{\textbf{Results using EMA of model parameters without English data}. We report validation perplexity (\textit{PPL}), average downstream accuracy (\textit{Dwn}) and Copain accuracy (\textit{Cop}).}
\label{tab:ema}
\end{table}

\subsection{EMA of Model Parameters}
\label{subsec:ema}

As discussed in \S\ref{subsec:param-shift}, including English significantly
reduces the parameter shift during CPT, which can presumably explain why this
variant is less prone to catastrophic forgetting. In this section, we explore
taking the EMA of the parameters \cite{morales2024exponential, cha2021swad} as
an alternative approach to reduce the parameter shift without requiring any
English data.


More concretely, every $\eta$ steps EMA sets the model parameters to a weighted
average between their current value and their value $\eta$ steps ago:
\[
  \resizebox{1\columnwidth}{!}{$
      \theta_t =
      \begin{cases}
        \theta'_t                                                      & \text{if } t \leq 0 \lor t \bmod \eta \neq 0 \\
        \alpha \theta_{t - \eta} + \left( 1 - \alpha \right) \theta'_t & \text{otherwise}                             \\
      \end{cases}
    $}
\]
where $\theta'_t$ and $\theta_t$ denote the model parameters at step $t$ before
and after EMA is applied, respectively, and $\alpha$ denotes the decay rate,
which we set to 0.92 in all of our experiments. Unless otherwise indicated, we
use $\eta=1$ for Basque and Indonesian, and $\eta=10$ for Arabic.

As shown in Table \ref{tab:ema}, EMA is competitive with conventional CPT
without the need for any English data. More concretely, it obtains the best
validation perplexity in all cases, and comparable results on downstream tasks.
This corroborates that the benefit of including English can be tied to
alleviating the parameter shift during CPT, and a similar effect can be
obtained by using EMA as a regularizer.

\begin{figure*}[t!]
\centering
\includegraphics[width=0.9\linewidth]{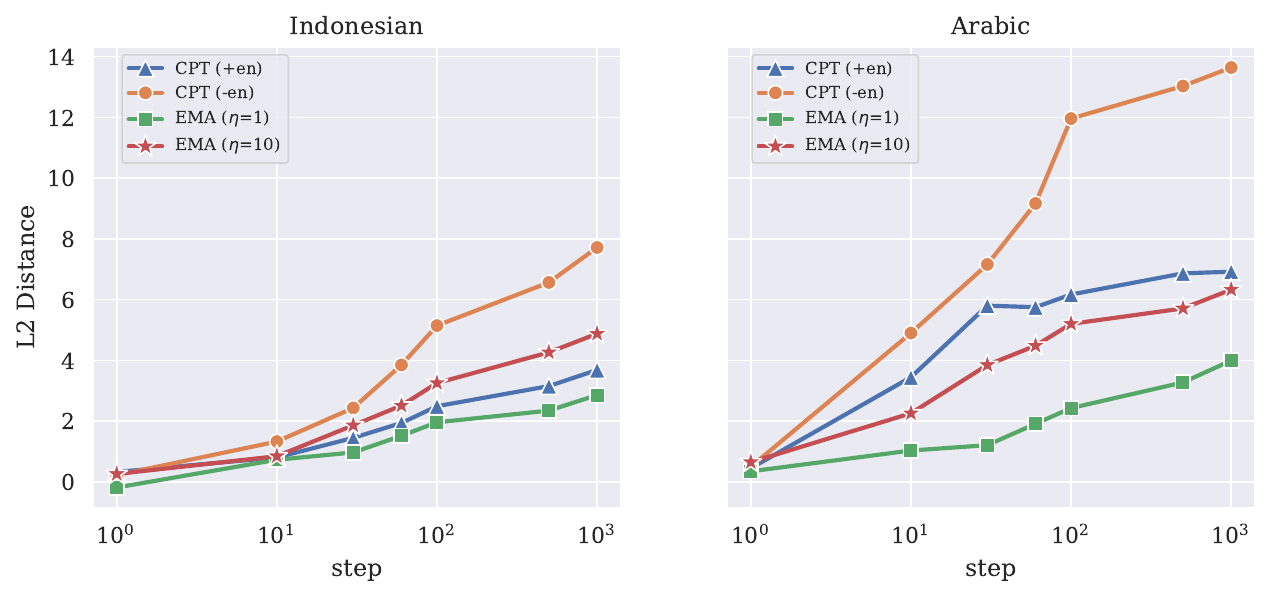 }
\caption{\textbf{L2 distance of model parameters from the initial Llama 2 (7B) during CPT} with and without English data, and using EMA with interval ($\eta$) of 1 and 10.}
\label{fig:l2norm-ar-id}
\end{figure*}

However, differences in Copain are bigger and more inconsistent. For instance,
EMA outperforms vanilla CPT by 10 points for Llama 2 7B in Arabic, but
underperforms it by 5 points for Llama 2 13B in Basque. In relation to this, we
found that the interval of applying EMA, $\eta$, had a big impact during our
preliminary experiments: \textbf{lower values result in more constrained
  parameter updates}, which helps mitigate the catastrophic forgetting of ICL
capabilities, but can potentially obstruct the learning of the target language.
Intuitively, we want to set $\eta$ so the shift in parameters is comparable to
that of vanilla CPT with English. But, as shown in Figure
\ref{fig:l2norm-ar-id}, this requires different values of $\eta$ depending on
the language. While outside the scope of this work, this prompts for future
work to develop more robust methods that can find a good trade-off without
excessive hyperparameter tuning.

\section{Related Work}

\paragraph{CPT for Language Adaptation.} The increased availability of LLMs lays a strong foundation for adapting models
to new domains through CPT \citep{abnar2021exploringlimitslargescale}. Lately,
CPT has been employed to expand LLMs to new languages or boost their
performance in languages where they previously struggled
\citep{gogoulou2024continuallearninglanguageshift}. Compared to training from
scratch, CPT achieves promising results by efficiently transferring the
knowledge and abilities learned by English-centric LLMs to target languages
\citep{fujii2024continualpretrainingcrosslingualllm}. Full-parameter CPT has
been shown to be efficient, provided that sufficient data is available in the
target language \citep{etxaniz2024latxaopenlanguagemodel,
  luukkonen-etal-2023-fingpt, yong2023bloom1addinglanguagesupport}. For
low-resource languages, LoRA is often leveraged in CPT. Separate LoRA weights
can be trained for each target language
\citep{fujii2024continualpretrainingcrosslingualllm,
  badola2023parameterefficientfinetuningrobustcontinual} or for all languages
collectively \citep{ji2024emma500enhancingmassivelymultilingual} and then
merged with the original weights. However, comprehensive investigations on the
effectiveness of LoRA in CPT are limited.

\paragraph{Stability Gap in Continual Learning.}

Continual learning aims to accumulate knowledge in deep neural models
\cite{parisi2019continual}. It is often used to extend pretrained models to new
domains, tasks, and languages. However, the ongoing distribution shift leads to
catastrophic forgetting of previous capabilities
\cite{ghunaim2023realtimeevaluationonlinecontinual}. A large body of literature
focused on mitigating this forgetting, mentioning it often occurs as a
transition phase to the new distribution
\citep{delange2023continualevaluationlifelonglearning,
  caccia2022newinsightsreducingabrupt}: the phase is referred to as the
\textit{Stability Gap}. During this gap, models lose performance on previously
learned tasks before recovering during training, or sometimes not at all. Our
analysis shows an analogous yet extreme case of the stability gap. Namely, we
see a rapid loss in ICL on Copain (\S\ref{sec:copain}), from which the model
struggles to recover.

\paragraph{EMA of Model Weights.}

EMA stabilizes the training of deep learning models. It is often employed in
approaches that focus on improving the generalization of the final model or
models close to convergence \cite{cha2021swad, yang2019swalp,
  izmailov2018averaging}. Furthermore, EMA allows the use of higher learning
rates, which is particularly beneficial for training LLMs with large effective
batch sizes \citep{morales2024exponential}. Lately, EMA gained wider use in
alignment of LLMs \citep{ouyang2022traininglanguagemodelsfollow}. For example,
it is used as a dynamic anchor in regularization to prevent forgetting of
pretrained knowledge while optimizing for rewards during Reinforcement Learning
from Human Feedback (RLHF) \citep{ramé2024warpbenefitsweightaveraged} and
Proximal Policy Optimization (PPO)
\citep{schulman2017proximalpolicyoptimizationalgorithms}.

\section{Conclusion}


In this paper, we have shown that including English data in CPT can be critical
for downstream capabilities to emerge in the new language, despite not having
an impact in validation perplexity. This can be traced back to a critical
period early on CPT, during which a drastic change in the training distribution
when switching to a new language causes a big shift in the model parameters,
which in turn results in the catastrophic forgetting of its ICL capabilities.
Based on these insights, we have shown that curriculum learning and EMA can
achieve the same effect while reducing---or fully eliminating---the need for
English data, further validating our findings.

While the focus of our work was to analyze the dynamics by which emergent
abilities arise during CPT, we believe that our insights can be helpful to
develop better strategies for language adaptation in the future. In particular,
one of our key findings is that controlling the degree of parameter shift is
critical for good downstream performance: giving too much flexibility can
result in the catastrophic forgetting of ICL, but overly constraining it can
hinder the learning of the target language. Our results with both LoRA and EMA
show that finding the right balance can be very sensitive to hyperparameters,
and even including English is not a universal solution as reflected by the
considerable drop in Copain performance in the case of Arabic and Indonesian
(Table \ref{tab:final-performance}). In the future, we want to explore more
robust CPT approaches that can find the optimal trade-off without the need for
excessive hyperparameter tuning.

\section*{Limitations}

Our analysis of emergent abilities was limited to multiple-choice downstream
tasks and language-agnostic ICL. It would be interesting to extend the study to
other capabilities, both from the perspective of language-independent skills
that might suffer from catastrophic forgetting, and target language skills that
may or may not emerge depending on the training dynamics. However, the scarcity
of relevant benchmarks, in particular in low-resource languages, hinders this
study.

In addition, our experiments were limited to including English in combination
with the target language. Experimenting with other high-resource languages
could provide additional insights, in particular when closely related to the
target language.


\section*{Acknowledgments}

This work has been partially supported by the Basque Government (Research group funding IT1805-22). 
Ahmed Elhady holds a PhD grant supported by the Basque Government (IKER-GAITU project). We also thank Aimar Zabala from the HiTZ Centre for his valuable inputs in some experiments.
The models were trained on the Leonardo supercomputer at CINECA under the EuroHPC Joint Undertaking, project EHPC-EXT-2024E01-042.

\bibliography{custom}


\appendix

\section{Initial Experiments}
\label{app:prel-exp}

In our initial experiments, we experimented with CPT of the Llama 2 7B model
with Basque data only using smaller learning rates to reduce the impact of
catastrophic forgetting. The results of these experiments are shown in Table
~\ref{tab:prel-exp}. We find that reducing the learning rate up to a factor of
10 not only did not solve the problem, but it also hindered the learning of the
new language.

\begin{table}[t!]
\centering
\small
\begin{tabular}{lrrrr}
\toprule
&\textbf{LR}&\textbf{PPL}&\textbf{Dwn.}&\textbf{Cop.}\\
\midrule
Llama-2 (7B)    &--&23.64&27.43&44.67\\
\quad + CPT (eu)&1e0-4&3.58&28.89&20.12 \\
\quad + CPT (eu)&5e0-5&8.76&27.98&26.43 \\
\quad + CPT (eu)&1e0-5&8.29&27.42&29.43 \\
\bottomrule
\end{tabular}
\caption{\textbf{Preliminarily experiments} to determine whether lower learning rates can reduce the impact of forgetting of the ICL capabilities during CPT. We use the Llama-2-7B model and do CPT with Basque data only.}
\label{tab:prel-exp}
\end{table}

\section{Continued Pre-training using LoRA}
\label{app:lora}
Table ~\ref{tab:lora} shows the results of using LoRA in CPT compared to full parameter CPT with and without including English data.

\begin{table}[t!]
\centering
\small
\begin{tabular}{lrrr}
\toprule
&\textbf{PPL}&\textbf{Dwn.}&\textbf{Cop.}\\
\midrule
Llama-2 (7B)    &23.64&27.43&\textbf{44.67}\\
\quad + CPT (eu+en)&\bfund{3.35}&\bfund{34.14}&43.43 \\
\quad + CPT (eu)&3.58&28.89&20.12 \\
\quad + LoRA (eu) & 3.68& 28.03 & 39.61\\ 
\bottomrule
\end{tabular}
\caption{\textbf{Results of CPT using LoRA} compared to full parameter CPT with and without English.}
\label{tab:lora}
\end{table}

\section{Results on Generative Benchmarks}
\label{res:mgsm-eu}
We evaluate the models on benchmarks that are natively curated in their respective languages. Unfortunately, no generative tasks were natively curated at the time of the writing. We only found the MGSM-eu benchmark \citep{baucells-etal-2025-iberobench} that evaluates mathematical reasoning in Basque. As shown in Table~\ref{tab:mgsm-eu}, the results are consistent with the findings in the MCQA benchmarks.

\begin{table}[t!]
  \small
  \centering
  \begin{tabular}{lc}
    \toprule
                            & \textbf{MGSM-Eu} \\
    \midrule
    \textbf{Llama 2 (7B)}   & 2.40             \\
    \quad + CPT (eu+en)     & 9.60             \\
    \quad + CPT (eu)        & 1.86             \\
    \quad + CPT w/ EMA (eu) & 5.60             \\
    \bottomrule
  \end{tabular}
  \caption{Performance of Llama-2-7B on the MGSM-eu benchmark using CoT. Accuracy is exact match (EM\%).}
  \label{tab:mgsm-eu}
\end{table}

\section{Multiple Choice Prompting Computation}
\label{app:mcp}
In multiple choice prompting, the model is prompted with few-shot demonstrations \textit{c} and a question \textit{q} and the set of choices $ A = \{A,B,C,D\}$. It generates a probability of the answer label $ a \epsilon A$ conditioned on the prefix prompt given by Equation ~\ref{eqn:icl-prob}. The model's answer is then set to:
\begin{equation}
  \renewcommand{\arraystretch}{0.4} %
  \label{eqn:ans}
  \argmax_{a\epsilon A}(P(a| c,q))
\end{equation}

\begin{table*}[t!]
\small
\centering
\begin{tabular}{lrrrrr}
\bottomrule
\rowcolor{gray!60}
\multicolumn{6}{l}{\textbf{Basque (eu)}}\\
\toprule
&\textbf{EusProf}&\textbf{EusExams}&\textbf{EusRead}&\textbf{EusTrivia}&\textbf{Average}\\
\midrule
\textbf{Random}&25.00&25.00&25.83&26.55&25.59\\
\midrule
\textbf{Llama 2 (7B)}&24.09&28.84&27.27&29.50&27.43\\
\quad + CPT (eu+en)&29.75&34.20&28.12&44.49&34.14\\
\quad + CPT (eu)&25.53&28.70&27.27&34.07&28.89\\
\quad + CPT (eu+en) (curr)&30.70&33.48&30.68&45.65&35.12\\
\quad + CPT w/ EMA (eu) &30.10&33.45&31.25&44.78&34.89\\
\quad + CPT w/ EMA (eu) (curr)&28.19&31.69&30.39&41.63&32.97\\
\quad + LoRA (eu) &25.82&27.90&28.49&29.93&28.03\\

\midrule
\textbf{Llama 2 (13B)} &25.90&29.66&28.98&33.53&29.52\\
\quad + CPT (eu+en)&41.73&40.05&36.09&52.22&42.52\\
\quad + CPT (eu)&33.35&32.08&28.69&46.70&35.20\\
\quad + CPT (eu+en) (curr)&41.92&40.19&35.39&52.18&42.42\\
\quad + CPT w/ EMA (eu) &40.80&40.30&32.67&51.77&40.39\\
\quad + CPT w/ EMA (eu) (curr)&40.80&40.30&32.67&51.77&40.39\\

\midrule
\textbf{Llama 3.1 (8B)}&32.52&48.01&43.03&45.70&42.31\\
\quad + CPT (eu+en)&53.34&54.55&60.47&54.67&55.75\\
\quad + CPT (eu)&52.54&53.41&59.07&53.33&54.84\\

\midrule
\textbf{Gemma 2 (9B)}&37.19&25.56&52.24&53.88&42.22\\
\quad + CPT (eu+en)&47.19&29.10&59.21&62.08&49.39\\
\quad + CPT (eu)&43.75&26.66&56.31&57.08&45.95\\

\bottomrule
\end{tabular}
\caption{Detailed downstream results on Basque}
\label{tab:detailed-perf-basque}
\end{table*}

\begin{table*}[t!]
\small
\centering
\begin{tabular}{lrrrrr}
\toprule
&\textbf{STEM}&\textbf{Humanities}&\textbf{Language}&\makecell[c]{\textbf{Social}\\ \textbf{Science}}&\makecell[c]{\textbf{Local}\\ \textbf{Culture}}\\
\bottomrule
\rowcolor{gray!60}
\multicolumn{6}{l}{\textbf{Arabic (ar)}}\\
\toprule
\textbf{Random}&29.50&28.60&25.80&28.90&32.30\\
\midrule
\textbf{Llama 2 (7B)} &33.70&32.65&28.40&32.80&34.70\\
\quad + CPT (ar+en)&35.02&35.23&32.36&33.82&35.31\\
\quad + CPT (ar)&34.40&35.18&28.24&31.73&33.79\\
\quad + CPT (ar+en) (curr)&37.23&34.51&32.35&31.78&36.76\\
\quad + CPT w/ EMA &34.48&33.22&31.52&31.34&36.23\\
\quad + CPT w/ EMA (curr)&33.21&31.95&29.02&30.13&34.45\\
\bottomrule
\rowcolor{gray!60}
\multicolumn{6}{l}{\textbf{Indonesian (id)}}\\
\toprule
\textbf{Random}&21.90&23.50&24.40&23.40&26.60\\
\midrule
\textbf{Llama 2 (7B)} &26.57&26.03&28.47&25.76&26.19\\
\quad + CPT (id+en)&28.78&30.85&32.33&31.92&30.19\\
\quad + CPT (id)&25.17&26.72&28.23&26.19&27.86\\
\quad + CPT (id+en) (curr)&28.64&28.73&30.97&27.39&29.72\\
\quad + CPT w/ EMA&28.45&27.67&31.40&27.84&30.21\\
\quad + CPT w/ EMA (curr)&27.64&26.37&28.97&26.97&27.38\\
\bottomrule
\end{tabular}
\caption{Detailed downstream results in Arabic and Indonesian}
\label{detailed-perf-ar-id}
\end{table*}

\begin{figure*}[t!]
\centering
\includegraphics[width=0.76\linewidth]{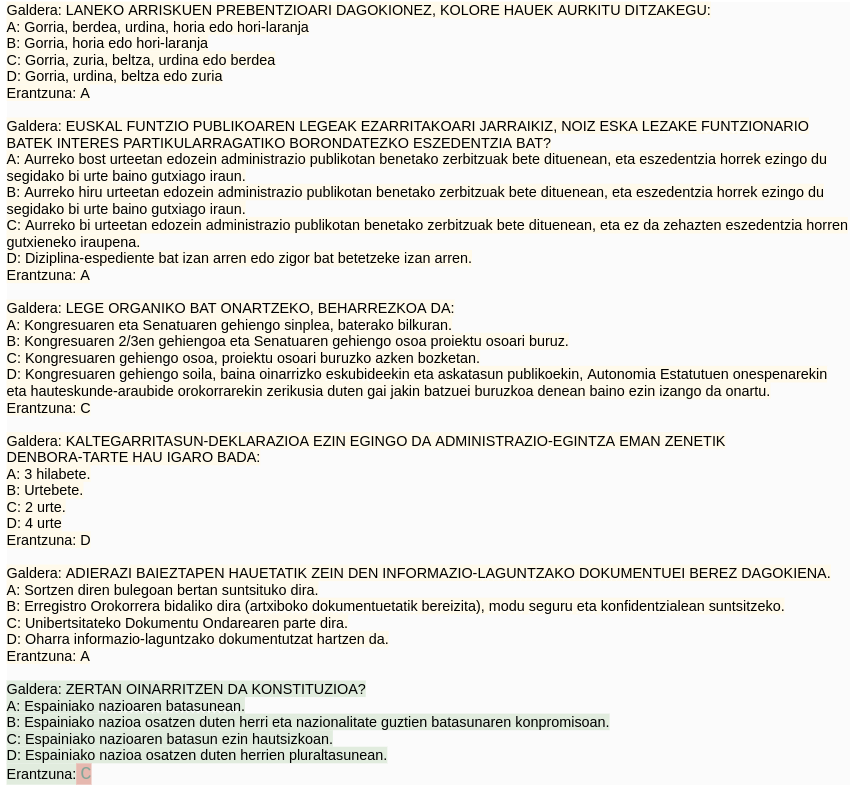 }
\includegraphics[width=0.76\linewidth]{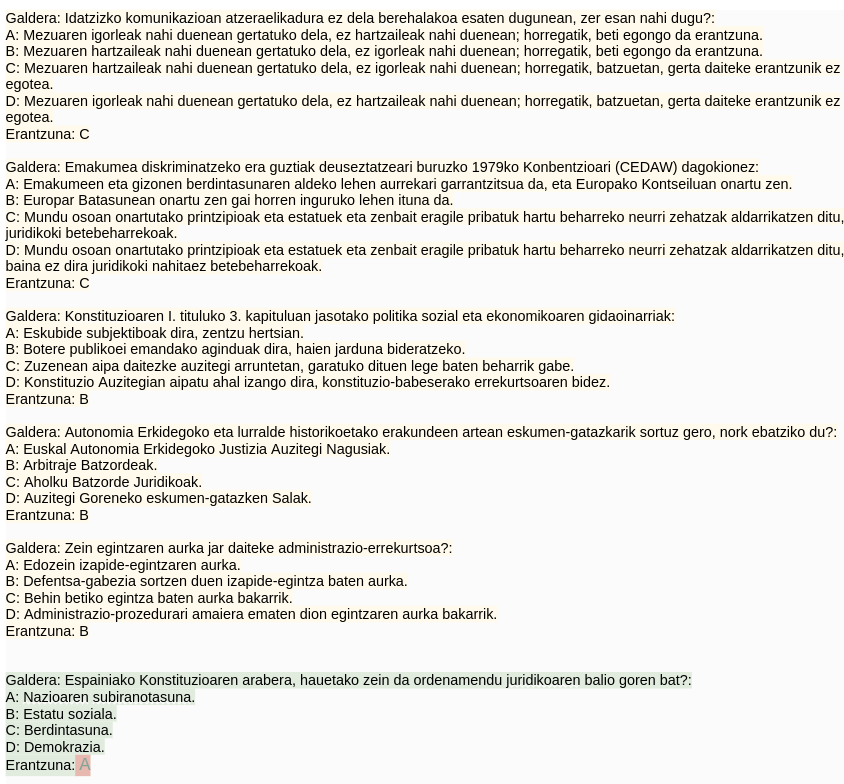 }
\caption{Example from EusExams using multiple choice prompting}
\label{fig:eusexams-1}
\end{figure*}

Figure ~\ref{fig:eusexams-1} shows examples from EusExams using multiple choice
prompting.

\section{Detailed Downstream Performance}

Table~\ref{tab:detailed-perf-basque} reports detailed downstream results in the
different subsets of Basque downstream tasks, while
Table~\ref{detailed-perf-ar-id} does so for Arabic and Indonesian.

\label{appx:full-res}




\end{document}